# Mean-Field Theory of Meta-Learning

*Dariusz Plewczynski* [1*]

[1] Interdisciplinary Centre for Mathematical and Computational Modelling, University of Warsaw, Pawinskiego 5a Street, 02-106 Warsaw, Poland, E-mail: darman@icm.edu.pl

[*] To whom correspondence should be addressed

**Keywords**

Mean-Field; Artificial Intelligence; Cellular automata; Minority survival; Computational Intelligence; Consensus Learning; Meta-Learning

**Abstract**

We discuss here the mean-field theory for a cellular automata model of meta-learning. The meta-learning is the process of combining outcomes of individual learning procedures in order to determine the final decision with higher accuracy than any single learning method. Our method is constructed from an ensemble of interacting, learning agents, that acquire and process incoming information using various types, or different versions of machine learning algorithms. The abstract learning space, where all agents are located, is constructed here using a fully connected model that couples all agents with random strength values. The cellular automata network simulates the higher level integration of information acquired from the independent learning trials. The final classification of incoming input data is therefore defined as the stationary state of the meta-learning system using simple majority rule, yet the minority clusters that share opposite classification outcome can be observed in the system. Therefore, the probability of selecting proper class for a given input data, can be estimated even without the prior knowledge of its affiliation. The fuzzy logic can be easily introduced into the system, even if learning agents are build from simple binary classification machine learning algorithms by calculating the percentage of agreeing agents.

## Introduction

The *Machine Learning* (ML) algorithms allow computers to learn based on training data. A major focus of ML is to recognize complex patterns in datasets, or make intelligent decisions based on data. Typically ML algorithms are divided into several classes: 1) supervised learning (generated a function that maps input data into desired outputs); 2) unsupervised learning (model a set of inputs, where no prior classification is given); 3) semi-supervised learning (generate an appropriate function or classifier); 4) reinforcement learning (learn how to act given an observation of the world, where every action has some impart in the environment, with feedback of it back to the algorithm); 5) transduction (predicts new outputs based on training inputs, outputs and test inputs); and 6) learning to learn (learns its own inductive bias based on previous experience) [1]. Different algorithms of ML have been applied successfully to solve real-life problems, for example in the context of bioinformatics [2-6] or chemoinformatics problems [7-12].

The current trend of machine learning theory and applications is to develop meta-learning techniques, since no single paradigm is superior to others in all possible situations [13-18]. What is the exact definition of *Meta-Learning*? Most ML researchers do not share any single definition of this term, therefore it is very important to precisely define what constitutes meta-learning in this manuscript. Traditionally this term refers to self-adaptation of ML methods, i.e. to readjusting their hypothesis spaces in order to fits better to changing environment and to allow for higher precision classification of incoming data, or making more accurate predictions about previously unknown cases. Yet, this very general statement cannot be easily described in terms of statistical mechanics and practical applications of such methodology. Therefore, I define here the meta-learning as the process of combining outcomes of individual learning procedures in order to determine the final decision. In that way, the individual learners perform typical machine learning procedure, then their predictions are gathered and integrated. This definition is very similar to other "meta" approaches in bio- and chemo-informatics, that are now of great interest in the field of computational biology and biophysics. My approach is currently well supported by recent advances in the protein fold recognition field that is dominated by the meta predictors like 3D-Jury [19, 20], Pcons [21, 22], Robetta [23-25]. Multiple tests confirmed that consensus methods were more powerful than individual prediction algorithm in sensitivity and specificity even if some meta-predictors used as little as three methods to build a consensus model. Yet, the problem of selecting the proper and most successful procedure of meta-learning is still ongoing debate.

Here, I provide a simple procedure for the integration of results from different methods into single prediction that complements previous approaches [14-18, 26]. As I mentioned above in the *Meta-Learning* (ML) one trains an ensemble of machine learning algorithms using different types of input training data representations [7, 27-29]. All possible solutions are gathered, and the consensus is build between them. The final phase of learning, i.e. consensus learning, is trying to balance the generality of solution and the overall performance of trained model. This approach is similar to other ensemble methods, yet differently from bagging (combines many unstable predictors to produce a ensemble stable predictor), or boosting (combines many weak but stable predictors to produce an ensemble strong predictor), it focuses of the use of heterogeneous set of algorithms in order to capture even remote, weak similarity of the predicted sample to the training cases. The main problem with such meta-approaches is that they are static and very specific. The meta-approach is optimized for certain combination of machine learning types of algorithms and selected particular representation of training data. Nevertheless, the model of the consensus should be in principle calculable in more general way.

Therefore, the goal of this manuscript is to provide a general, theoretical framework for the general integration of results individual machine learning algorithms. In order to perform analytical analysis, I assume infinite, statistical ensemble of different ML methods. The global preference toward true solution can be described in my approach as the global parameter affecting all learners. Each learner (intelligent agent) performs training on available input data toward classification pressure described by the set of positive and negative cases. When the query testing data is analyzed each agent predicts the query item classification by "yes"/"no" decision. The answers of all agents are then gathered and integrated into the single prediction via dynamical evolution described within cellular automata framework. This dynamical view of the consensus between various machine learning algorithms is especially useful for artificial intelligence, or robotic applications, where adaptive behavior given by the integration of results from ensemble of ML methods.

**Brainstorming: Cellular Automata model of Meta-Learning**

My model of learning is based on nonlocal *cellular automata* (CA) approach known from physics [30-34], with a wide range of applications [35-43]. The first statistical mechanics of opinion formation in groups of individuals was proposed by Lewenstein et al. [44] on the class of models that were based on probabilistic cellular automata and social impact theory introduced by Latane [45, 46]. The mean-field theory with intermittent behavior was observed with a variety of stationary states with a well-localized and dynamically stable clusters (domains) of individuals who share minority opinions [44]. The impact of a group of $N$ agents on a given learner is proportional to three factors: 1) the *"strength"* of the members of the whole ensemble, 2) their *"social"* distance from the individual, and 3) their number $N$. Such model leads to ferromagnetic and spin-glass phases, when different values of persuasiveness and supportiveness are assumed. Than this approach was successfully used in a variety of sociological phenomena, with an interesting extension of the model was done by Kohring [47, 48], where Latane's theory was extended to include learning. The cellular automata with intrinsic disorder was later solved analytically in the continuous limit by Plewczynski [49], and proved that in the model of Cartesian social space (therefore not fully connected) and containing no learning rules, one can also observe different phases (small clusters in the sparse phase with large role of strong individuals, and high density phase with almost uniform opinion). The later results of Hołyst et al., where numerical simulations and analytical models were tested in simplified geometries, proved the usefulness of mean-field formalism in describing the social impact theory, and the presence of the equilibrium states of the system with complex intermittent behavior [50-53].

In the present manuscript I present a novel application of cellular automata models to the meta-learning problem. Each cell in CA represent single machine learning algorithm, or certain combination of parameters, and optimization conditions affecting the classification output of this particular method. I call this single learner by the term "learning **agent**", each characterized for example by its prediction quality on a selected training dataset. The coupling of individual learners is described by short-, medium- or long-range interaction strength, so called learning **coupling**. The actual structure, or topology of coupling between various learners is described using term "**learning space**", and can have different representations, such as Cartesian space, fully connected or hierarchical geometry. The result of the evolution, dynamics of such system given by its stationary state is defined here as the consensus equilibration. The majority of learners define, in the stationary limit, the "learning **consensus"** outcome of the meta-learning procedure. For example by the total difference between positive and negative predictions in the binary classification problem.

The information integration, i.e. the consensus building between various machine learning algorithms, various prediction outcomes, is similar to the dynamical changes in cellular automata systems known from physics. The phase transitions can be observed in the system, the global new phase emerging when the system reaches a critical point in terms of its order parameter. Changes between phases of the system are induced by some external factors that can be modeled as a bias added to the local fields. Mean field theory was successfully used in a plenty of physical problems, such as a super-fluid effects [54], weakly interacting Bose gas in external field [55-59], quantum solitons in optical fibers [60, 61] and many others [62-68]. The main difference between my approach and those formalisms is the procedure of taking real function instead of complex one. The individual learners are modeled here, as single machine learning procedures, therefore each of them is described by set of calculable real values depending on the each ML method's precision, recall and classification error, therefore assuming the input training data. My approach is supervised learning because the description of the quality for each agent depends on the type or representation of the training data. On the other hand, the presented formalism is more general allowing for unsupervised learning, i.e. for searching for unknown patterns in input data, by representing them on the fully connected grid, and allowing for features similarity exchange. This approach is now studied by Plewczynski et al. in the context of two wide range applications in bioinformatics: namely the protein-protein interactions prediction and protein-ligand docking. In the first case, the whole known proteome of an organism is represented on the two dimensional grid, where each node represent single protein. The coupling between two proteins is given by the structural and sequence similarity focused on likely interactions sites on the surfaces of proteins [69]. This approach allows for unsupervised protein-protein interactions prediction, taking different route, namely unsupervised meta-learning based on cellular automata and phase transitions than previous methods [70-74].

The cellular automata model of meta-learning is based on several assumptions:

1. *Binary Logic*

   I assume the binary logic of individual learners, i.e. we deal with cellular automata consisting of *N* agents, each holding one of two opposite states ("NO" or "YES"). These states are binary $\sigma_i = \pm 1$, similarly to Ising model of ferromagnet. In most cases the machine learning algorithms that can model those agents, such as support vector machines, decision trees, trend vectors, artificial neural networks, random forest, predict two classes for incoming data, based on previous experience in the form of trained models. The prediction of an agent answers single question: is a query data contained in class A (“YES”), or it is different from items gathered in this class (“NO”).

2. *Disorder and random strength parameter*

   Each learner is characterized by two random parameters: persuasiveness $p_i$ and supportiveness $s_i$ that describe how individual agent interact with others. Persuasiveness describes how effectively the individual state of agent is propagated to neighboring agents, whereas supportiveness represent self-supportiveness of single agent. In present work I assume that influential agents has high self-esteem $p_i = s_i$, what is supported by the fact that highly effective learners should have high impact on others in meta-learning procedure. For example, we can select $p_i = f(precision, i)$ and $s_i = f(recall, i)$ in the case where agents are modeled as single machine learning procedures. In general the individual differences between agents are described as random variables with a probability density $\hat{p} = (p_i, s_i)$, with mean values $p = \sum \frac{p_i}{N}$ and $s = \sum \frac{s_i}{N}$. Similarly to the social influence theory, the quality of predictor in some way affect its influence strength, when the final optimization of meta-learning consensus is done.

   In the case of meta-learning procedure the persuasiveness $p_j$ represents here the ability of learning agent *j* to persuade agents who hold the opposite state to switch to having the same state as *j*. The supportiveness $s_j$ represents the ability of learning agent *j* to support agents who hold the same state, so not only the self-support of an individual agent (to itself), but the support that an agent gives to other agents who share the same state as it has.

3. *Learning space and learning metric*

   Each agent is characterized by a location in the learning space, therefore one can calculate the abstract learning distance $d(i, j)$ of two learners *i* and *j*. The strength of coupling between two agents tend to decrease with the learning distance between them.

Determination of the learning metric is a separate problem, and the particular form of the metric and the learning distance function should be empirically determined, and in principle can be a very peculiar geometry. In present manuscript, I select the fully connected learning space, where all distances between agents are equal $d(i,j)=1$. This particular geometry is useful for example in the case of simple consensus between different yet not organized machine learning algorithms, where no group of learners perform significantly better than others.

4. *Learning coupling*

   Agents exchange their opinions by biasing others toward their own classification outcome. This influence can be described by the total learning impact $I_i$ that *i*th agent is experiencing from all other learners. Within the cellular automata approach this impact is the difference between positive coupling of those agents that hold identical classification outcome, relative to negative influence of those who share opposite state, and can be formalized as

$$I_i = I_p\left(\sum_j \frac{p_j}{N}\left(1-\sigma_i\sigma_j\right)\right) - I_s\left(\sum_j \frac{s_j}{N}\left(1+\sigma_i\sigma_j\right)\right), \tag{1}$$

   where $I_p(.)$ and $I_s(.)$ are the functions of persuasiveness and supportiveness impact of the other agents on the *i*-th agent. It should be noted here that the persuasiveness $p_j$ represents here the ability of agent *j* to persuade agents who hold the opposite state to switch to having the same state as *j*. On the contrary the supportiveness $s_j$ represents the ability of agent *j* to support agents who hold the same state, i.e. preventing them from switching to the opposite state. That is, persuasiveness represents the propensity of *j* to cause other agents to switch to her state, and supportiveness represents her propensity to keep them there.

5. The equations of *meta-learning*

   The equation of dynamics of the learning model defines the state $\sigma_i^{'}$ of *i*th individual at the next time step as follows:

$$\sigma_i^{'} = \left(-sign\left(\sigma_i I_i\right)\right), \tag{2}$$

   with rescaled learning influence:

$$I_i = \sum_j \frac{p_j}{N(s+p)}(1-\sigma_i\sigma_j) - \sum_j \frac{s_j}{N(s+p)}(1+\sigma_i\sigma_j). \quad (3)$$

I assume a synchronous dynamics, i.e. states of all agents are updated in parallel. In comparison to standard Monte Carlo methods the synchronous dynamics takes shorter time to equilibrate than serial methods, yet it can be trapped into periodic asymptotic states with oscillations between neighboring agents.

6. Presence of *noise*

   The randomness of state change (phenomenological modeling of various random elements in the learning system, and training data) is given by introducing noise into dynamics:

$$\sigma_i^{'} = \left(-sign\left(\sigma_i I_i + h_i\right)\right), \quad (4)$$

   where $h_i$ is the site-dependent white noise, or one can select a uniform white noise, where for all agents $h_i = h$. In the first case $h_i$ are random variables independent for different agents and time instants, whereas in the second case $h$ are independent for different time instants. I assume here, that the probability distribution of $h_i$ is both site and time independent, i.e. it has uniform statistical properties. The uniform white noise simulates the global bias affecting all agents, whereas site-dependent white noise describes local effects, such as prediction quality of individual learner etc.

The system defined in this way is similar to previously postulated cellular automata models of opinion change in social sciences [44, 49, 50]. The main differences of those approaches from the previously described cellular automata models is given by the infinite-range interactions and fully connected cellular automata, that are better fitted to the learning context of the problem. In addition, the random strength parameters are introduced, therefore allowing for more complex behavior to be observed. Individual agents are described using probability density $\hat{p} = (p_i, s_i)$, so they differs from each other. The impact function is also included, so learners are able to exchange their states in the form of coupling.

**Mean-Field approximation**

The fully connected learning space geometry present an interesting practical formalizations for further analysis of meta-learning procedure. Here, all agents are coupled with each other with some randomly distributed strength that is independent on the distance between them. The mean-field theory provides very well defined and controlled approximation allowing for solving the dynamical equations of such model. The dynamical "order" parameter has to be defined, to show the decay of minority groups in the form of "staircase" dynamics. The fully connected geometry of learning space is supported by topology of this problem. We have a set of independent machine learning algorithms, no prior hierarchy or topology for this set is imposed, and we would like to build the consensus between their predictions.

The discrete equation of dynamic is given by:

$$\sigma_i^{'} = -sign\left( \sum_j \frac{p_j}{(s+p)N} \left(\sigma_i - \sigma_j\right) - \sum_j \frac{s_j}{(s+p)N} \left(\sigma_i + \sigma_j\right) + h_i \right), \tag{5}$$

Introducing a weighted majority-minority difference for a system:

$$m = \sum_j \frac{\left(s_j + p_j\right)\sigma_j}{N\left(s+p\right)}, \tag{6}$$

and random parameters to describe effective self-supportiveness of each agent:

$$a_i = \frac{s-p}{s+p} + \frac{\beta}{s+p} s_i, \tag{7}$$

we get the dynamical equation in noise absent limit by rewriting the Eq. 5 using $\theta(.)$ as Heaviside theta function:

$$\sigma_i^{'} = sign\left(m+h_i\right)\theta\left(\left|m+h_i\right| - \left|a_i\right|\right) + \sigma_i sign\left(a_i\right)\theta\left(\left|a_i\right| - \left|m+h_i\right|\right). \tag{8}$$

We assume here that $a_i \geq 0$ for any distribution of random variable $s_i$ [44].

The order parameter is defined in physics as a quantity that defines the phase transition between various phases in the physical system, for example defining the evolution of the ordered system toward chaotic behavior. In this model, the order parameter describes the changes between various meta-learning solutions (agreement between learners, uniformity of opinion, minority clusters, chaotic state of not coupled learners) and it is given by the formula:

$$\pi\left(\varphi\right) = \sum_j \frac{\left(s_j + p_j\right)\sigma_j}{N\left(s+p\right)} \theta\left(a_j - \varphi\right), \tag{9}$$

as in the standard mean-field theory of Ising systems [44, 47, 62]. The meaning of this parameter is as follows: $\pi$ is a positive real number, and it is equal to the weighted majority-minority difference calculated for those agents that have effective self-supportiveness $a_j$ greater that $\varphi$. The order parameter defines different, stationary states of the dynamics, in the noiseless limit it determines uniquely the approach toward a stationary state, and with small noise the only stationary states are close to uniformity with $m \cong \pm 1$. From the practical point of view it is enough to describe the ordering among agents of a given strength in order to completely specify the state of the system.

The derivative $\frac{\partial \pi(\varphi)}{\partial \varphi}$ is related to the weighted majority-minority difference for agents with effective self-supportiveness $a_j$ is equal to $\varphi$, i.e. for those agents with $s_j = \frac{1}{\beta}\left[\varphi(s+p)-(s-p)\right]$. As in the work of Lewenstein [44] the order parameter $\pi_i$ in the noiseless limit fulfills the equation:

$$\pi^{'}(\varphi) = \left[\mu(m,\varphi) + \pi(|m|)\right]\theta(|m|-\varphi) + \pi(\varphi)\theta(\varphi-|m|), \tag{10}$$

with $\pi(0) = m$, and

$$\mu(m,\varphi) = {}^{1}\!/_{N}\, sign(m)\sum_j \frac{(s_j+p_j)\sigma_j}{N(s+p)}\theta(|m|-|a_j|)\theta(a_j-\varphi). \tag{11}$$

The mean-field approximation for the system is introduced by replacing the actual variables (like $m$, $\pi(\varphi)$, $\mu(m,\varphi)$ by their corresponding mean values calculated by averaging over disorder, i.e. random distribution of self-supportiveness $s_i$ and $p_i$. The averaged equations are then valid for very large *N* (preferably infinite system), where $m$ is no longer random variable, because its fluctuations are of order of ${}^{1}\!/_{\sqrt{N}}$. The recurrence equation for $m$ is defined for $\varphi = 0$, i.e. in regions where $\varphi \geq |m| = \pi(0)$:

$$m^{'} = \mu(m,0) + \pi(|m|). \tag{12}$$

Reformulating the definition of $\pi(\varphi)$ gives:

$$\pi(\varphi) = m - \sum_j \frac{(s_j+p_j)\sigma_j}{N(s+p)}\theta(\varphi - a_j). \tag{13}$$

Therefore the recurrence equation for $m$ has the following form [44]:

$$m' = m + \sum_j \frac{\left(s_j + p_j\right)\sigma_j}{N\left(s+p\right)}\left[sign\left(m\right) - \sigma_j\right]\theta\left(\varphi - a_j\right). \tag{14}$$

Let's define the function $f\left(m\right)$ as the value of $m$ in the consecutive time steps:

$$f\left(m\right) = m_0 + \sum_j \frac{\left(s_j + p_j\right)\sigma_j}{N\left(s+p\right)}\left[sign\left(m\right) - \sigma_j(t=0)\right]\theta\left(\left|m\right| - a_j\right). \tag{15}$$

Therefore, we have the mean-field behavior of the recurrence equation described by the general equation:

$$f\left(m\right) = m_0 + \sum_j \frac{\left(s_j + p_j\right)\sigma_j}{N\left(s+p\right)}\left[sign\left(m\right) - \sigma_j(t=0)\right]\theta\left(\varphi - a_j\right). \tag{16}$$

In this meta-learning model each agent is described by two random parameters: persuasiveness $p_i$ (the coupling with other agents) and supportiveness $s_i$ (self-supportiveness). The first parameter, persuasiveness, imposes how effectively the individual state of agent is propagated to neighboring agents, whereas the second parameter, supportiveness, represents self-influence of a single agent. We can select for each type of machine learning algorithm, or instance of single ML method (depending for example on the methods' parameters values) a $p_i = f\left(precision, i\right)$ and $s_i = f\left(recall, i\right)$. In present manuscript I assume that influential agents has high self-esteem:

$$p_i = s_i = h\left(precision, recall, i\right) \approx \frac{1}{2}\left(precision\left[i\right] + recall\left[i\right]\right). \tag{17}$$

Here, highly effective learners should have high impact on others in meta-learning procedure. The individual differences between agents are described as random variables with a probability density $\hat{p} = \left(p_i, s_i\right)$, with mean values $p = \sum \frac{p_i}{N}$ and $s = \sum \frac{s_i}{N}$, therefore the recurrence equation (16) has a simplified form:

$$f\left(m\right) = m_0 + \sum_j \frac{s_j}{Ns}\left[sign\left(m\right) - \sigma_j(t=0)\right]\theta\left(\varphi - a_j\right), \tag{18}$$

with

$$a_j = \frac{\beta}{2s} s_j .$$

The procedure of averaging is done here over different random distributions of initial conditions $\sigma_i\left(t=0\right)$, and different possible distributions of values of self-supportiveness $s_i$.

The value of $m$ changes during the evolution of the system:

$$m = \sum_j \frac{s_j \sigma_j}{Ns},$$

and

$$\pi\left(\varphi\right) = \sum_j \frac{s_j \sigma_j}{Ns} \theta\left(a_j - \varphi\right).$$

The recurrence equation for $m^{'}$ is given by the formula [44]:

$$m^{'} = m_0 + \sum_j \frac{s_j}{Ns}\left[sign\left(m\right) - \sigma_j(t=0)\right]\theta\left(\left|m\right| - a_j\right).$$

And for positive initial value of $m$ we have:

$$m^{'} = \mu\left(m,0\right) + \pi_0\left(m\right).$$

Reformulating the definition of $\pi\left(\varphi\right)$ gives:

$$\pi_0\left(\varphi\right) = m - \sum_j \frac{s_j \sigma_j\left(t=0\right)}{Ns} \theta\left(\varphi - a_j\right), \tag{19}$$

therefore the recurrence map for $m$ has the following form [44]:

$$f\left(m\right) = m_0 + \sum_j \frac{s_j}{Ns}\left[1 - \sigma_j\left(t=0\right)\right]\theta\left(m - a_j\right). \tag{20}$$

The above equations show that $f\left(m\right)$ is bounded and increasing function of $m$. The $m$ is bounded and increasing function of time step, therefore it has at least one stable fixed point. It may have also several fixed stable points separated by unstable ones, fulfilling the equation:

$$\sum_j \frac{s_j}{Ns}\left[1 - \sigma_j\left(t=0\right)\right]\theta\left(m - a_j\right) = 0. \tag{21}$$

During time evolution $m$ tend to the nearest stable fixed point from its initial value $m_0$, therefore if noise is not present the meta-learning gives the final answer as the local minima in the space of possible solutions. Therefore clusters of minorities appear in the system as generic solutions in the noiseless limit. When the system is forced to make a consensus prediction for example by majority rule, the final prediction outcome is given simply by the majority result, yet the actual probability of the correct answer can be approximated as the percentage of states in agreement with majority rule, in comparison to the number of minority groups.

For large enough $m$ the system is reaches the stationary state close to the uniformity state. For small $m$ the system is unstable, and grows quadraticaly in $m$. The similar description is valid for negative values of $m$, therefore $f(m)$ has two stable and one unstable fixed point [44]. In the field theoretical formulation the system has a set of local minima describing clusters of minority states characterized by different $s_j$. Those groups collapse successively with stronger $s_j$ when a small noise is added to the system [44]. I have to point out, that above model is valid also for different initial conditions, where weaker agents have random state in opposition to uniform state of stronger ones. If the equilibrium state of the system exists, and it is stable, the system will be also stable for cases, where initial state for weaker agents will be more diverse (non-uniform).

The above solutions for the system are given by the minority clusters surrounded by the majority agents, and the dynamic is of "staircase" character in the presence of small noise [44, 49]. In general three initial states of the system can observed: sparse (no correlation between the agent strength and its state), middle density (a state of a agent starts to be correlated with its strength, lot of interesting meta-stable global configurations), and large density state (most of agents initially have similar state, therefore the role of coupling is not so important). In the first case clusters of both types of states may appear, and when the weak coupling is present there is no bias toward uniform solution. In the second case a variety of sophisticated geometries, shapes of clusters are present, some are robust and meta-stable, other disappearing slowly changing their state in agreement with majority rule. Here, no analytical solutions are easy to find, therefore computer simulations have to be applied. I leave this case to my next manuscript, where extensive computer modeling of the system will be presented.

**Concluding remarks**

Intelligent agents theory is a fascinating topic in modern science [75-79]. Decision making transitions depend to high degree on global factors influencing an ensemble of independent learners. On the other hand, those changes are dependent to a high degree on individual decisions (predictions) that are based on agents' attititudes. During consensus, i.e. the final decision making, the reciprocal influence is critical as each learner exchange its opinion with others. In my approach, I assume that external factors acting on each learner are present during only the first phase of meta-learning, where initial states for a population of learners are setting up. Yet, both processes even if acting on different time scales, are important for understanding the computational intelligence process.

In this manuscript I have presented the statistical theory of meta-learning. In my approach I select long-range coupling between agents, as opposite for example to the Euclidean two dimensional learning space, where only nearest-neighbors are coupled. This assumption is well supported by the fact that we are typically focused on only equilibrium, stationary states. The fully connected learning space lets agents evolve faster in comparison to other types of cellular automata. In addition, all agents influence each other, therefore we avoid local minima traps for the global system.

Each learner is characterized by two random parameters: persuasiveness $p_i$ and supportiveness $s_i$ that describe how individual agent interact with others. The random strength parameters simulate different individual features of learning agents. In principle one can define both parameters in various different ways. In the case of a set of machine learning algorithms, each of them can be described by its intrinsic parameters affecting precision of single classification model of training data. In general case, several different types of machine learning algorithms can be used as individual learners. There, the distribution of quality of local prediction can be described as random providing that algorithms differ significantly between each other in terms both of the quality of prediction (classification accuracy), recall values (the ability to memorize the positive items in the training dataset), or precision (the ability to precisely predict the classification of training items).

The other definition of those parameters (persuasiveness and supportiveness) can enhance the method persuasiveness (the value of $p_i$ ), if the method has the state $\sigma_i = +1$, and make its $p_i$ value lower when the opposite state is taken. In this way, it allows to speed up the consensus process by forcing system to reach equilibrium state more rapidly, yet pushing it to the +1 decision based on the selected training dataset. This can cause several problems with overtraining, therefore some limitations of this approach should be taken into account. The actual solutions presented in this paper, yet do not depend strongly on the selected form of those parameters. Anyway we assume that they are some random variables describing the variety of individual decisions in the ensemble of learners.

There two time scales in the system. The first time scale is related to the fast evolution of individual learners. When input testing data is presented to the system, each learner respond by its own single prediction. This local prediction of each agent is done very rapidly, almost instantly. Then those individual predictions are processed by cellular automata algorithm in order to find the stationary state of the system. This part is denoted as integration of information. As it was shown above, such stationary state has the form of minority clusters surrounded by the sea of majority prediction. Therefore, the final consensus prediction given by the majority rule, still preserves non-orthodox solutions, allowing for fast adaptivity of the system when training data pattern is changed. The time scale for this integrative process is relatively long in comparison to individual predictions, therefore very fast (preferably optimized for parallel processing) cellular automata software implementations have to be prepared in order to apply described above formalism in real life problems. In the statistical model presented here, I assume that there is no coupling between those two time scales. Therefore I neglect all details of individual evolution of learners, focusing our attention for integration phase of incoming local information into single, consensus answer.

The core question of this manuscript is how typical initial distribution of learners' state evolve in time? As it was shown above different initial conditions distinguish by the numbers of agents sharing opposite opinion $m = \sum_j \sigma_j / N$ into three classes. The first class is close to uniform state $|m| \cong 1$, where almost all learners initially are in agreement. The evolution of the system rapidly collapses into stationary uniform state. This situation is observed when individual learners share similar machine learning algorithms, or wide spectra of parameters values do not change the classification model. Opposite states are sparse, and randomly spread over the learning space. For example, most of single ML algorithms (such as Random Forest, SVM) trained on an easy or with moderate difficulty training dataset will give very similar prediction for a test cases. Therefore the initial state of the consensus system is close to uniformity of opinion, and the uniformity state is the most frequent final state. The second initial condition $0 \ll |m| < 1$ describes much richer solutions space, moderate number of agents share opposite state and those can be distributed randomly over the learning space, or clustered into well defined groups. This describe the situation, where different values of parameters can cause different classification outcomes, or ensemble of ML algorithms contain significantly different between each other algorithms, that construct distinct classification models of input training data. The system has its intrinsic preferences (or in other words preferable local classification model) – most of agents agree with their preferences, yet to some degree the opposite consensus state is possible. Therefore, one can assume that agreement between agents is possible, even if there is significant proportion of learners that classify input data oppositely. The third type of initial conditions $|m| \cong 0$ contain distributed randomly or clustered different agents' states spread over the learning space. Because the number of opposite states is similar, therefore the system is on the edge of phase transition between two final consensus answers: "YES" or "NO". Therefore, even small perturbation of initial state, parameters change or type and nature of testing examples can in principle guide the system into different, opposite answers. This type of consensus is more fitted to the difficult training cases, where both answers are very probable. Here, the final trained system is very fragile and strongly depend on testing input data. The small change of input testing data can build up the very different consensus value. Here, the consensus as the final, stable state of the whole system is not obvious, and it can take a significant amount of time. The final state can be either randomly distributed negative learners in the majority of positive states, or clustered minorities.

Here, I apply the analytical results of mean-field approximation for cellular automata dynamics with moderate- and long-range interactions known previously from findings of Lewenstein et al. [44], Plewczynski [49] and Hołyst et al. [50, 51, 53] to novel problem of meta-learning. The CA model is characterized by long-range interactions between individual agents, the fully connected learning space (each individual learners exchange their prediction results with all others) and by an intrinsic disorder that allow for complex learning geometries to appear in the system. Two obvious emergent phenomena were observed immediately in this class of models: polarization and clustering [44]. In this article I prefer to use different terms for description of those emerging phenomena, namely *integration* and *adaptivity*. The integration is based on polarization effect (single majority as the stable stationary state of the system, therefore allowing for consensus prediction), whereas the adaptivity is given by the clustering of similar outputs of groups of learners (small minority clusters of non-preferred learning outcome are grouped together in the learning space).

The order parameter of the system is given by the variable $\eta$ that characterizes geometrical and dynamical features of the model. The equilibration of the system in clustered state is given by intermittent, consecutive steps in the form of "stair-case" dynamics [44, 49]. First the strongest agents change their state, then the weaker rest of minority cluster collapses. The final solution is given in most cases as single state with uniform prediction outcome. Therefore is defines the final answer of the consensus system, when input data is presented to the learning network of individual agents. The existence of this uniform solution in the limit of infinite time, when noise is present, is of crucial importance for further analysis of this meta-learning model. The real life realizations of these algorithm will be presented in forthcoming manuscripts, especially in the field of bio- and chemo-informatics.

**Acknowledgements:**

This work was supported by the Polish Ministry of Education and Science (N301 159735) and other financial sources. I would like to thank Prof. M. Lewenstein (ICREA & ICFO, Barcelona, Spain) and Prof. M. Niezgodka (ICM, University of Warsaw, Warsaw, Poland) for stimulating discussions. Author would like to thank anonymous reviewer for fruitful comments and suggestions that strongly enhanced the results of this work.